\documentclass[runningheads]{llncs}
\usepackage{graphicx}
\usepackage{amsmath,amssymb,amsfonts}%

\usepackage{subcaption}

\usepackage{rotating}
\usepackage{booktabs} 
\usepackage{float}
\begin{document}
\title{Optimizing Deep Reinforcement Learning for Adaptive Robotic Arm Control}
%

\author{Jonaid Shianifar\inst{1}\orcidID{0000-0003-0477-0056} \and
Michael Schukat\inst{1} \and
Karl Mason\inst{1}}
\authorrunning{J. Shianifar et al.}
%
\institute{School of Computer Science, University of Galway, H91 FYH2, Galway, Ireland 
\email{\{J.Shianifar1,Michael.Schukat,Karl.Mason\}@universityofgalway.ie}}
\maketitle              
\begin{abstract}

In this paper, we explore the optimization of hyperparameters for the Soft Actor-Critic (SAC) and Proximal Policy Optimization (PPO) algorithms using the Tree-structured Parzen Estimator (TPE) in the context of robotic arm control with seven Degrees of Freedom (DOF). Our results demonstrate a significant enhancement in algorithm performance, TPE improves the success rate of SAC by 10.48 percentage points and PPO by 34.28 percentage points, where models trained for 50K episodes. Furthermore, TPE enables PPO to converge to a reward within 95\% of the maximum reward 76\% faster than without TPE, which translates to about 40K fewer episodes of training required for optimal performance. Also, this improvement for SAC is 80\% faster than without TPE.  This study underscores the impact of advanced hyperparameter optimization on the efficiency and success of deep reinforcement learning algorithms in complex robotic tasks.

\end{abstract}

\section{Introduction}
In the rapidly evolving field of robotics, the development of autonomous and highly adaptable robotic arms represents a significant step towards realizing more efficient, precise, and versatile automated systems \cite{arif2022mix}. These systems find applications across a spectrum of industries, from intricate tasks in surgical robotics to the repetitive, precision-demanding processes in manufacturing \cite{liu2021deep}. The control of robotic arms, especially those with a high DOF, poses a substantial challenge due to the complexity of their operation and the need for fine-tuned coordination \cite{imran2022open,horgan2023evolving}. A promising answer to these challenges is DRL, bringing a framework for robotic arms to learn and optimize their activity through interaction with the environment. \cite{liu2021deep,han2023survey,kober2013reinforcement}. DRL methods have demonstrated their efficiency in solving a wide spectrum of tasks—from games to the control of robotic systems \cite{zhang2021traded,he2022robust,dong2023enhanced,pan2023robotic}. 

Among the various algorithms under the DRL paradigm, SAC and PPO have stood out for their effectiveness in balancing exploration and exploitation, which is critical for the efficient learning of control policies in environments with high-dimensional action spaces. SAC, known for its off-policy learning and entropy regularization, encourages efficient exploration and has demonstrated success in continuous control tasks \cite{haarnoja2018soft}. On the other hand, PPO, an on-policy algorithm, simplifies the implementation and has shown robust performance across a range of DRL problems \cite{schulman2017proximal}. However, the optimal application of these algorithms in robotic arm control, particularly for arms with 7-DOF, heavily depends on the precise tuning of their hyperparameters.

The tuning process, which is often manual and intuitive, can be significantly enhanced using optimization algorithms. The TPE emerges as a potent tool in this aspect, offering a structured and efficient means of hyperparameter optimization. By employing a Bayesian model approach the relationship between hyperparameters and the objective function, TPE facilitates the identification of optimal configurations with fewer evaluations \cite{bergstra2011algorithms}.

This study is the first to apply TPE optimization to DRL algorithms, SAC and PPO, for robotic arm control, filling a significant gap in research on hyperparameter impact on DRL performance.
Also, This paper makes the following contributions:
\begin{enumerate}
    \item To optimize the hyperparameters of two DRL algorithms, SAC and PPO, utilizing the TPE for enhanced control of a robotic arm with 7-DOF.
    \item To demonstrate through a series of experiments the effectiveness of TPE in significantly improving the precision, convergence speed, and overall task success rate of the SAC and PPO algorithms in a reach target task.
    \item To highlight the importance of SAC and PPO hyperparameter optimization in achieving performance in robotic arm control. 
\end{enumerate}
The paper is organized as follows: Section 2 details the methodologies, including DRL frameworks, hyperparameter optimization, the defined task, and the experimental setup involving the Franka Emika Panda arm in a simulated environment. Section 3 discusses the results, demonstrating the effectiveness of the TPE in enhancing the performance of algorithms for robotic arm control. The final section presents the conclusions and future work.

\section{Materials and Methods}

\subsection{Deep Reinforcement Learning}
\begin{figure}[h]
\centering
\includegraphics[width=0.5\textwidth]{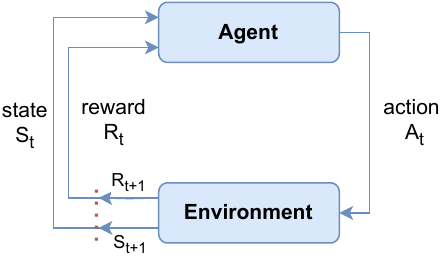}
\caption{In DRL, the agent selects actions based on its policy, interacts with the environment, and receives rewards to adjust its policy for future actions.} \label{fig-RL}
\end{figure}

DRL's application in robotics has been marked by its ability to tackle complex, high-dimensional control tasks, a domain where traditional programming methods fall short \cite{lindner2021positioning,lobbezoo2023simulated,imran2022open}. DRL is a pivotal area within machine learning that centers on teaching agents to make strategic decisions through interactions with their environment, aiming to maximize rewards. Within the DRL framework, an agent systematically engages with its environment. At each timestep $t$, the agent observes a current state $s_t$, selects an action $a_t$ based on its policy $\pi$, and receives a corresponding reward $r_t$, thereby transitioning to the next state $s_{t+1}$. This iterative process persists until reaching a terminal state, indicating the end of an episode. The agent's overarching aim is to devise a policy that optimally increases the aggregate expected reward over time, The process of DRL is depicted in Fig.~\ref{fig-RL}. DRL aligns well with the autonomous decision-making required in robotics \cite{kober2013reinforcement,arulkumaran2017deep}. 

Recent advancements in DRL have further expanded its capabilities, integrating deep neural networks to approximate complex functions and policies \cite{mnih2015human}. We explore two prominent algorithms in this context, PPO and SAC.

\subsubsection{Proximal Policy Optimization (PPO)}
is a popular DRL algorithm known for its stability and sample efficiency. It aims to optimize the policy while ensuring that the updates are not too drastic \cite{schulman2017proximal}. The objective function for PPO can be expressed as:

\begin{equation}
    L^{CLIP}(\theta) = \hat{\mathbb{E}}_t \left[ \min(r_t(\theta) \hat{A}_t, \text{clip}(r_t(\theta), 1-\epsilon, 1+\epsilon) \hat{A}_t) \right]
\end{equation}

Here, $L^{CLIP}(\theta)$ is the clipped objective function, $\hat{\mathbb{E}}_t$ denotes the expected value over a finite batch of samples, $r_t(\theta) = \frac{\pi_\theta(a_t|s_t)}{\pi_{\theta_{old}}(a_t|s_t)}$ is the probability ratio of the current policy $\pi_\theta$ over the old policy $\pi_{\theta_{old}}$, $\hat{A}_t$ is an estimator of the advantage function at time $t$, and $\epsilon$ is the clipping parameter \cite{schulman2017proximal}.
Table~\ref{tab_ppo} lists several key hyperparameters essential for tuning, highlighting the numerous hyperparameters associated with PPO.
\begin{table}
\caption{Hyperparameters for Proximal Policy Optimization (PPO).}
\label{tab_ppo}
\begin{tabular}{@{}lp{8.5cm}c@{}}
\toprule
\textbf{Name} & \textbf{Description} & \textbf{Symbol}  \\
\midrule
learning\_rate & Learning rate of the optimizer & \(\alpha\) \\
n\_steps & Number of steps to collect data for each iteration & -  \\
batch\_size & The batch size used for updating the policy network & -  \\
gamma & The discount factor, which determines the importance of future rewards & \(\gamma\)  \\
ent\_coef &The coefficient for the entropy term in the loss function, encouraging exploration  & \(\alpha\)  \\
vf\_coef & The coefficient for the value function loss, which determines the balance between policy and value function updates  & -  \\
max\_grad\_norm & The maximum gradient norm, which can be used to clip gradients during optimization to prevent exploding gradients & - \\
gae\_lambda & The generalized advantage estimation (GAE) lambda parameter, which controls the trade-off between bias and variance in the advantage estimate & \(\lambda\) \\
clip\_range & The clipping range for the surrogate loss, which helps in preventing&- \\
\hline
\end{tabular}

\end{table}

\subsubsection{Soft Actor-Critic (SAC)}
 is a DRL algorithm that combines off-policy and actor-critic methods \cite{haarnoja2018soft}. It is particularly suited for continuous action spaces. The SAC objective includes maximizing the entropy of the policy to encourage exploration. The SAC objective is given by:

\begin{equation}
J(\theta) = \mathbb{E}_{\tau \sim \pi_{\theta}} \left[ \sum_{t=0}^{\infty} \gamma^t \left(r(s_t, a_t) + \alpha H(\pi(\cdot|s_t))\right)\right]
\end{equation}

Here, $J(\theta)$ represents the SAC objective, $\theta$ are the policy parameters, $\tau$ is a trajectory sampled from the policy, $r(s_t, a_t)$ is the immediate reward, $\alpha$ is the temperature parameter, a positive scalar that multiplies the entropy term in the objective function, and $H(\pi(\cdot|s_t))$ is the entropy of the policy \cite{haarnoja2018soft}. 
The SAC algorithm involves multiple hyperparameters, as Table~\ref{tab_SAC} outlines. Similar to PPO, the efficacy and efficiency of SAC in generating optimal policy decisions are considerably dependent on the precise adjustment of its hyperparameters.
\begin{table}
\caption{Hyperparameters for Soft Actor-Critic (SAC).}
\centering
\label{tab_SAC}
\begin{tabular}{@{}lp{8cm}c@{}}
\toprule
\textbf{Name} & \textbf{Description} & \textbf{Symbol} \\
\midrule
buffer\_size & The size of the replay buffer used for storing past experiences  & - \\
learning\_starts & Number of steps before learning begins & -  \\
batch\_size & The batch size used for sampling from the replay buffer during training & -  \\
tau & The soft update coefficient, which controls the rate at which the target networks are updated & \(\tau\)  \\
gamma & The discount factor, which determines the importance of future rewards & \(\gamma\) \\
learning\_rate & Learning rate for the optimizer & \(\alpha\)  \\
ent\_coef & The coefficient for the entropy regularization term in the policy loss  & \(\alpha\)  \\
target\_update\_interval & Interval for target network updates & -  \\
gradient\_steps & Number of gradient steps per update & -  \\
use\_sde & Whether to use state-dependent exploration & -  \\
\hline
\end{tabular}

\end{table}

\subsection{Tree-structured Parzen Estimator (TPE)}
TPE, as named by Bergstra et al. \cite{bergstra2013hyperopt}, is the technique of using Bayesian optimization heuristics for guiding and speeding up the hyperparameter configuration optimization process. Its innovative approach involves the creation of a binary tree-like model that adeptly maps out the probability distributions for various hyperparameters. This model is particularly adept at navigating the complex landscapes often encountered in high-dimensional spaces, where objective functions can be costly to evaluate, thereby streamlining the optimization trajectory towards optimal solutions with efficiency and precision \cite{bergstra2013hyperopt}.
Mathematically, TPE optimizes by iteratively selecting hyperparameters based on the following principle:
\begin{equation}
\theta^* = \arg\min_\theta \frac{P(\text{Objective better than current best} \mid \theta)}{P(\text{Objective worse than current best} \mid \theta)}
\end{equation}

where $\theta$ represents the optimized hyperparameters, and $P(|)$ denotes the conditional probabilities that the objective function is better or worse than the current best given the chosen hyperparameters. This Bayesian approach enables TPE to efficiently explore the search space by balancing exploration and exploitation, ultimately guiding our evolutionary algorithm towards optimal solutions.

\subsection{Task Definition}
The reach target task, pivotal in robot manipulation, necessitates the end effector's precise targeting within Cartesian space constraints across diverse tasks \cite{han2023survey}. The targets are randomly generated within this space to assess the arm's ability to adapt and accurately reach different points, simulating potential real-world applications \cite{gallouedec2021panda}. Our objective is to optimize DRL for training a 7-DOF robotic arm to reach target tasks, where a policy learns the mapping from current states to subsequent actions for goal achievement. Notably, existing literature predominantly addresses position-only reaching. This study delineates the reach task by explicitly defining state, action, goal, and reward \cite{gallouedec2021panda}.

\begin{itemize}
    \item \textbf{State ($S_t$)}: Represents the environment's current status, comprises a vector that includes joint angles ($\theta$), the end-effector position($EE_{pos}$), and the goal position ($Goal_{pos}$).
    \item \textbf{Action ($A_t$)}: The action corresponds to the individual motion of each joint, an action provided by the agent to update the environment state.
    \item \textbf{Reward ($R_t$)}: As a reward function, we use a dense function, shown on Equation~.\ref{eq_reward}, where the closer the agent is to completing the task, the higher the reward.
\begin{equation}
    \label{eq_reward}
    R_{t} = - \sqrt{(x_{EE} - x_{Goal})^2 + (y_{EE} - y_{Goal})^2 + (z_{EE} - z_{Goal})^2}
\end{equation}
    \item \textbf{Done}: A binary flag indicating the end of an episode, task is completed when the distance between the end-effector and the goal position is less than 5 cm.
    \begin{itemize}
        \item \textbf{Terminated}: Episode ends due to goal achievement.
        \item \textbf{Truncated}: Episode ends due to reaching a step limit, not because the task was completed, reaching the step limit in the training phase is 50, and for the testing phase is 5 steps.
    \end{itemize}
    
\end{itemize}

\subsection{Training and Evaluation}

\begin{figure}[h]
    \centering
    \includegraphics[width=.72\linewidth]{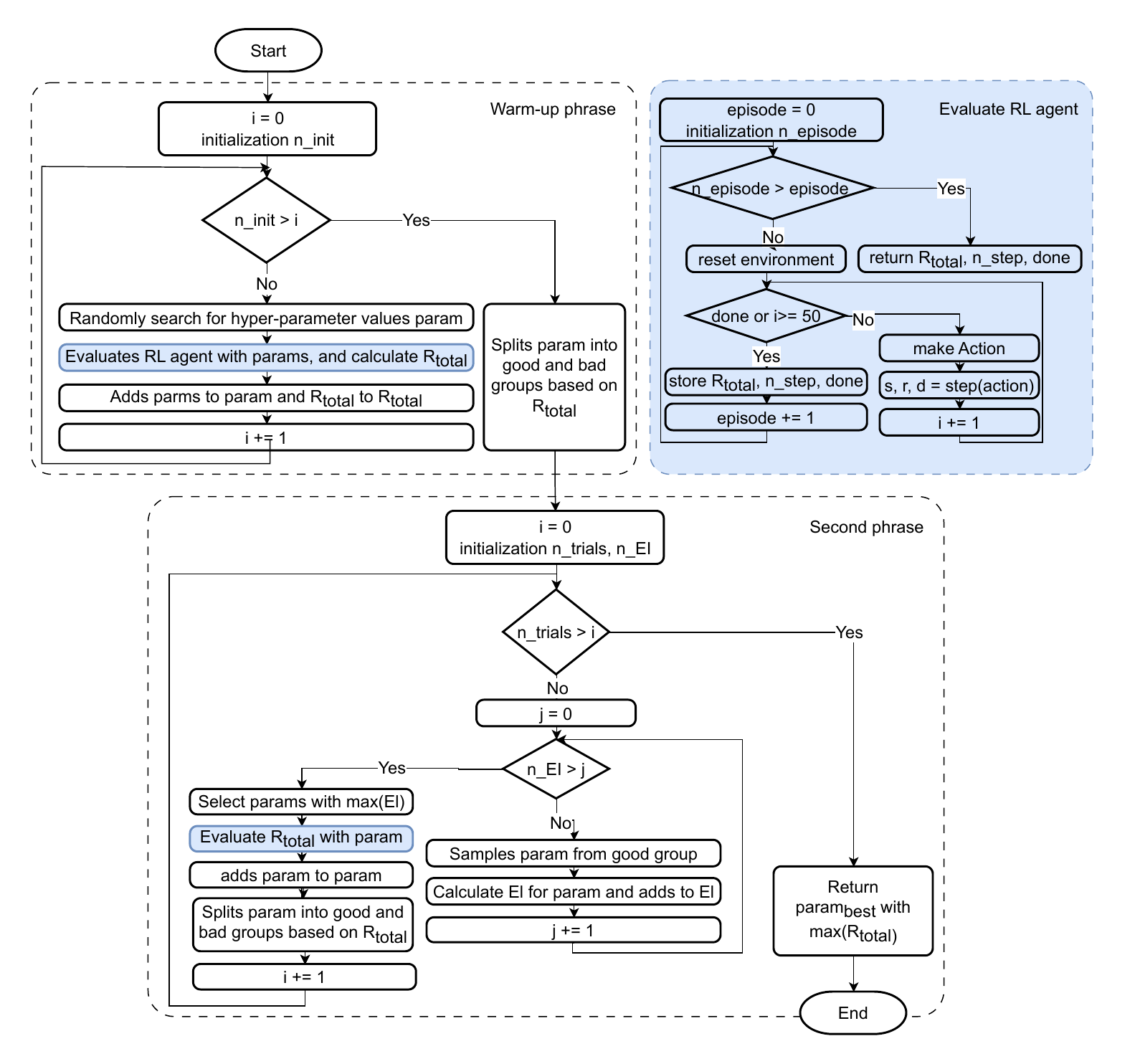}
    \caption{The iterative process of hyperparameter optimization and the DRL model evaluation.}
    \label{fig_training}
\end{figure}

The methodology used in this work is depicted in Fig.~\ref{fig_training}. The process encompasses two primary phases: a warm-up phase for initial hyperparameter exploration and a subsequent phase for focused training and hyperparameter refinement.

\subsubsection{A. Training Process Overview}
The training commences with an initial phase dedicated to exploring a range of hyperparameters through random sampling, aiming to identify a foundational set that yields a favorable reward outcome. This stage, referred to as the warm-up, sets the groundwork for more targeted optimization.

Following the initial exploration, the training enters a critical phase where the algorithm focuses on refining the model's accuracy and responsiveness. Actions are evaluated based on their success in navigating towards a set goal under the constraints of the environment, with each episode offering a fresh challenge. The objective is to systematically improve the model's performance through iterative learning, with a specific cap on the number of steps (50 steps per episode) to encourage efficiency and deter protracted decision-making paths.

Finally, the models trained with hyperparameters selected with TPE and also default hyperparameters, for 100,000 epochs, where the robot must find a random target in a maximum of 50 steps.

\subsubsection{B. Evaluation Methodology}
Post-training, the model is tested against 100,000 randomly generated target positions. The evaluation emphasizes not only the success rate in reaching these targets, but also the efficiency as measured by the number of steps required, imposing a stricter limit of 5 steps per episode during this phase. This sharpens the focus on models ability to quickly adapt and accurately respond to new scenarios. Also, models saved in 20K, and 50K training episodes are tested to show learning convergence speed.

\subsection{Experimental Setup}
\begin{figure}[h]
\centering
\begin{subfigure}{.5\textwidth}
  \centering
  \includegraphics[width=.8\linewidth]{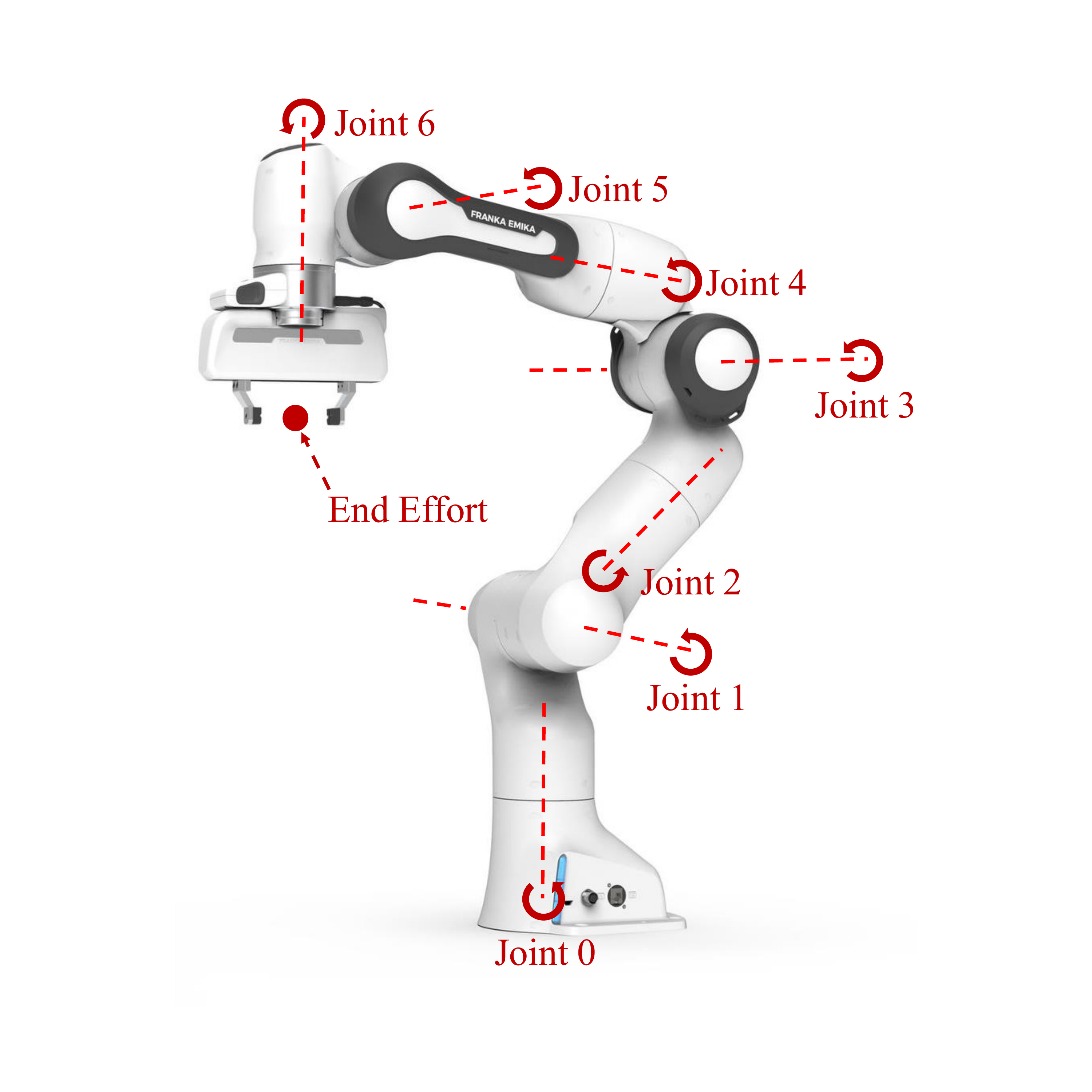}
  \caption{}
  \label{fig:sub1}
\end{subfigure}%
\begin{subfigure}{.5\textwidth}
  \centering
  \includegraphics[width=.8\linewidth]{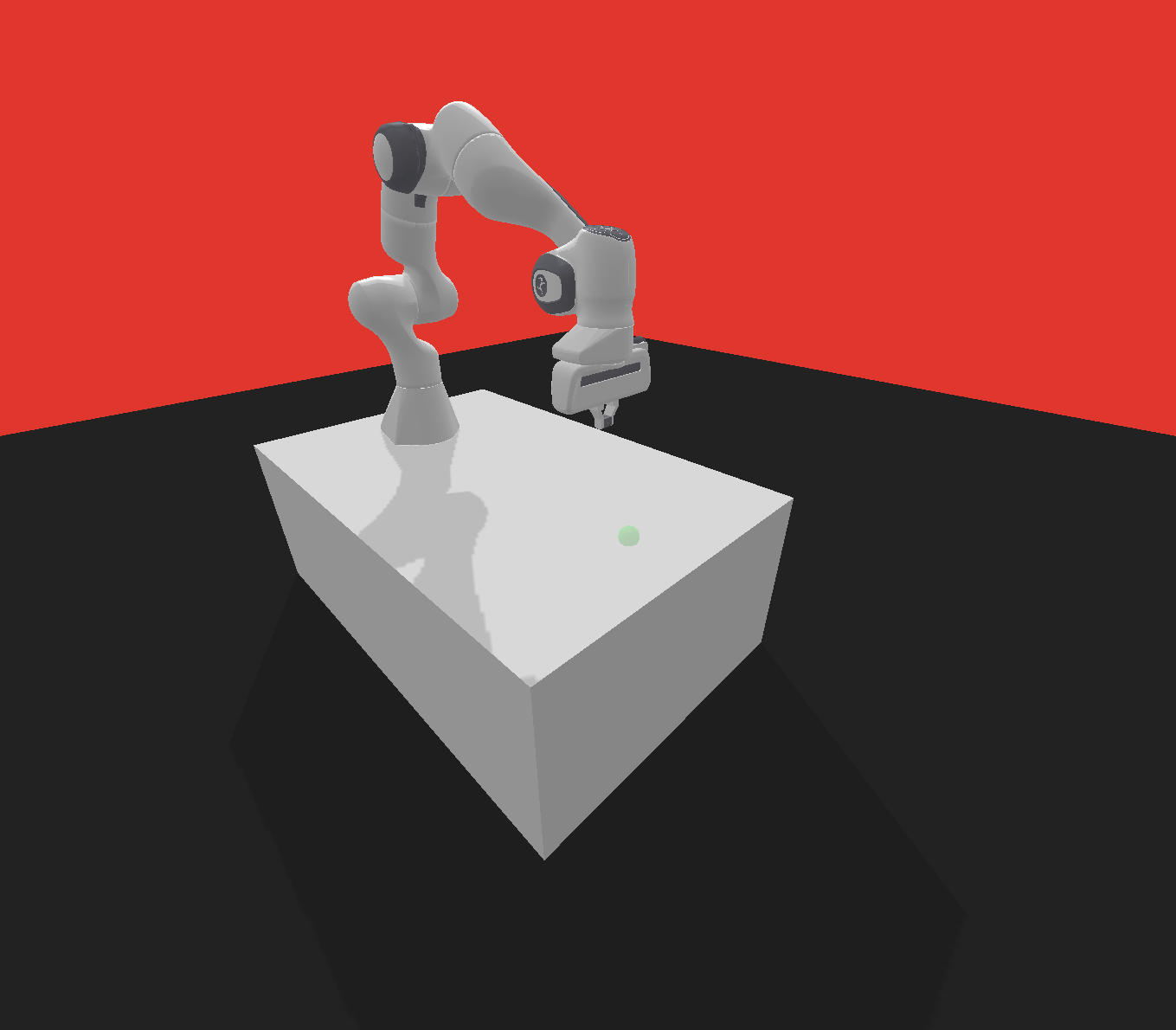}
  \caption{}
  \label{fig:sub2}
\end{subfigure}
\caption{(a) Franka Emika Panda arm robot with declared joints position, (b) panda\_gym simulation environment}
\label{figpandarobot}
\end{figure}

This study utilized the Franka Emika Panda arm with 7-DOF in a panda\_gym \cite{gallouedec2021panda} simulation, developed with PyBullet \cite{coumans2021} and Gymnasium \cite{towers_gymnasium_2023}, to conduct risk-free tests (Fig.~\ref{figpandarobot}). Our experiments ran on a system with a 12th Gen Intel Core i7-12700 CPU and 64GB of RAM, using Python 3.8.10 for programming. We applied PPO and SAC via the Stable Baseline3 \cite{stable-baselines3} library and optimized hyperparameters with the Optuna \cite{akiba2019optuna} framework.

\section{Results}
Our evaluation focused on the performance of two DRL algorithms, SAC and PPO, in conjunction with TPE for hyperparameter optimization, aiming to enhance the control of a robotic arm with 7-DOF. We observed significant improvements in learning efficiency, the utility of TPE in hyperparameter selection, and insights from various data visualizations and analyses. A key metric for evaluation was the success rate in reaching targets across the testing phases, which highlighted the enhanced learning efficiency and the strategic optimization of hyperparameters facilitated by TPE.

\subsection{Hyperparameter Optimization}
\begin{figure}[h]
    \begin{minipage}{0.5\textwidth}
        \centering
        \includegraphics[width=\linewidth]{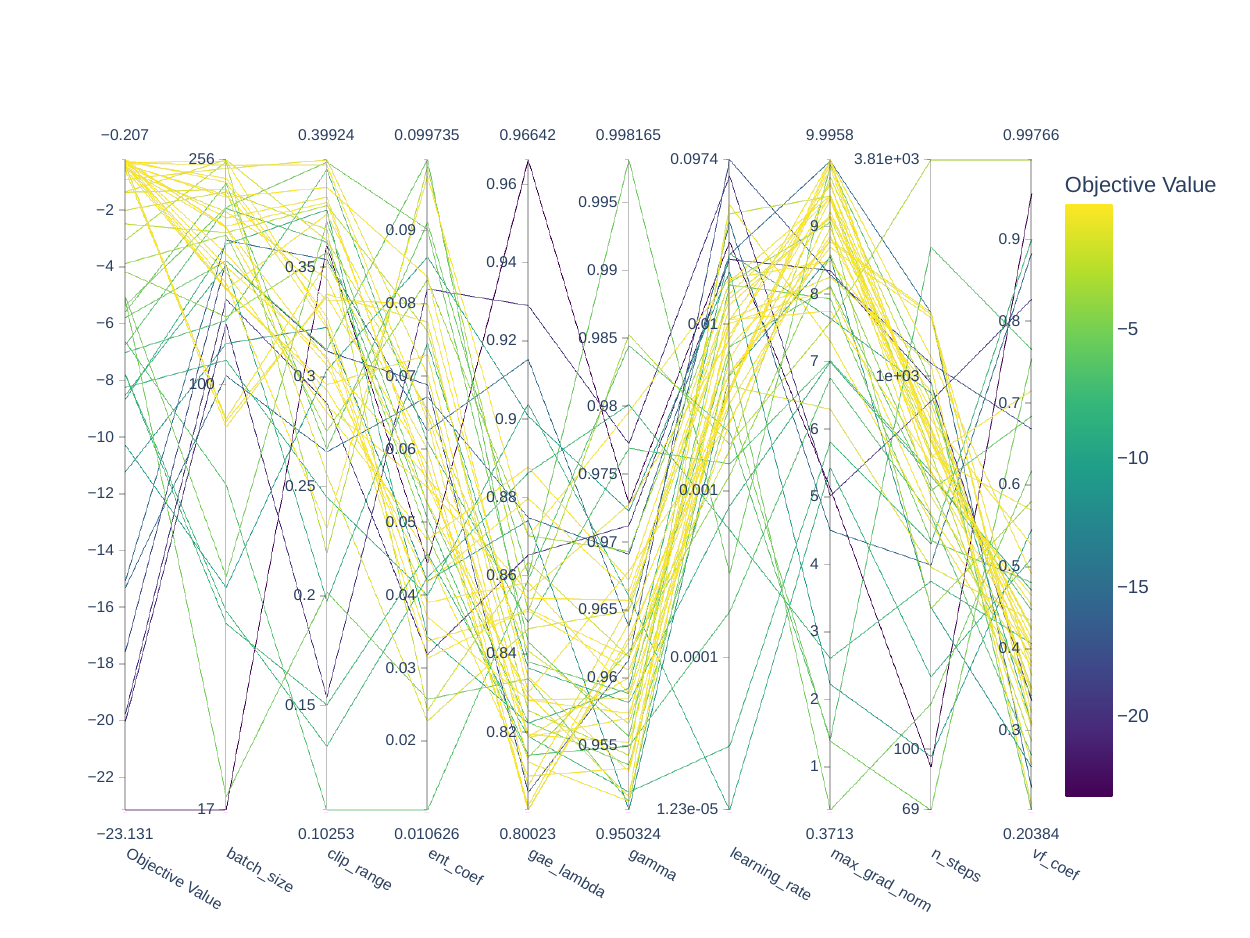}
        
    \end{minipage}%
    \begin{minipage}{0.5\textwidth}
        \centering
        \includegraphics[width=0.9\linewidth]{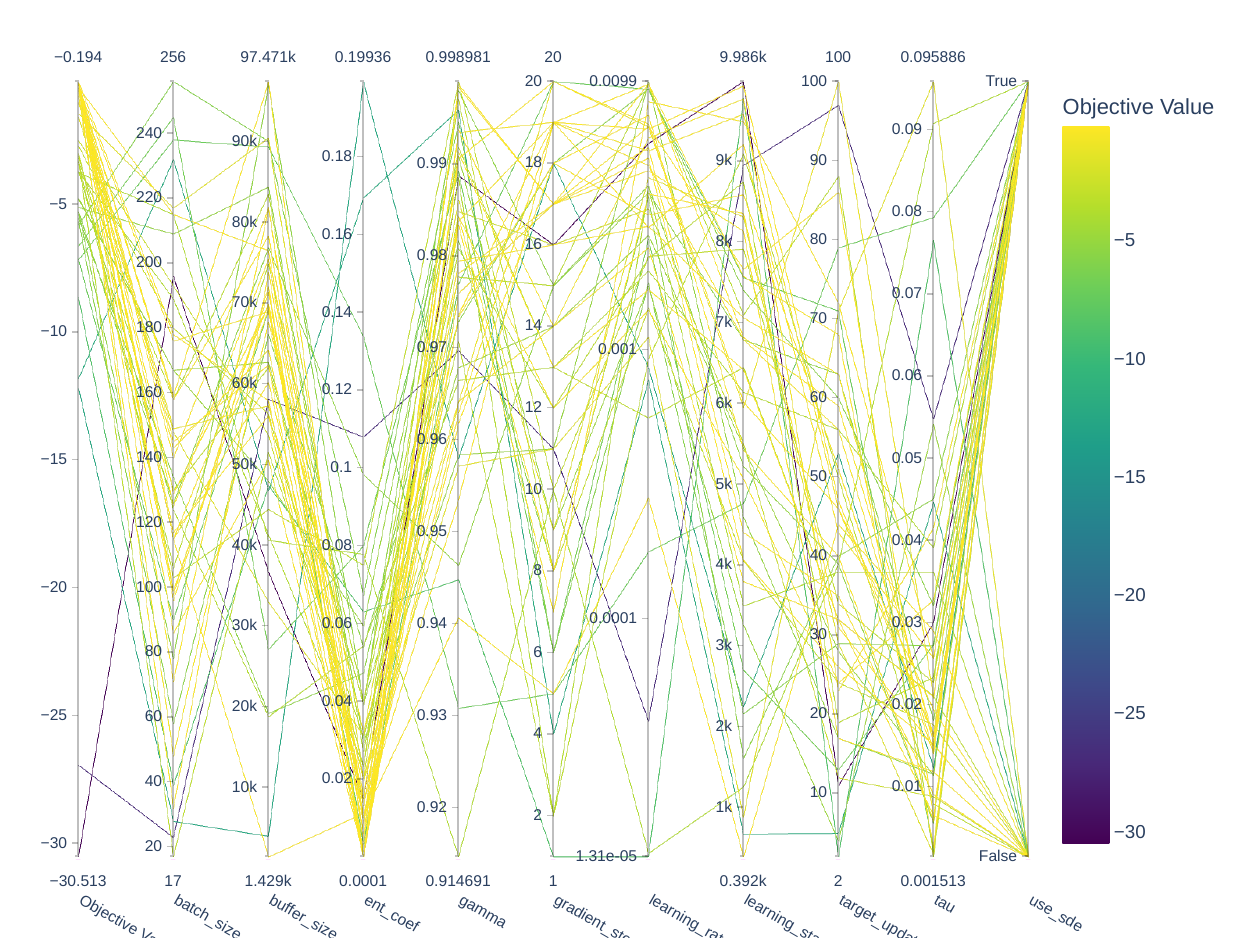}
        
    \end{minipage}
    \caption{Parallel coordinate plot(PCP) for PPO \textbf{(left)} and SAC \textbf{(right)}.}\label{pcp}
\end{figure}

The optimization procedure benefited significantly from the strategic selection of hyperparameters via TPE, as shown in Fig.~\ref{pcp}, parallel coordinate plots (PCP) for PPO and SAC elucidated the systematic exploration within the hyperparameter space, charting the trajectories toward optimized performance. This comprehensive view demonstrated how various hyperparameter configurations contributed to the overall efficacy of the learning models.

\begin{figure}[h]
    \begin{minipage}{0.5\textwidth}
        \centering
        \includegraphics[width=\linewidth]{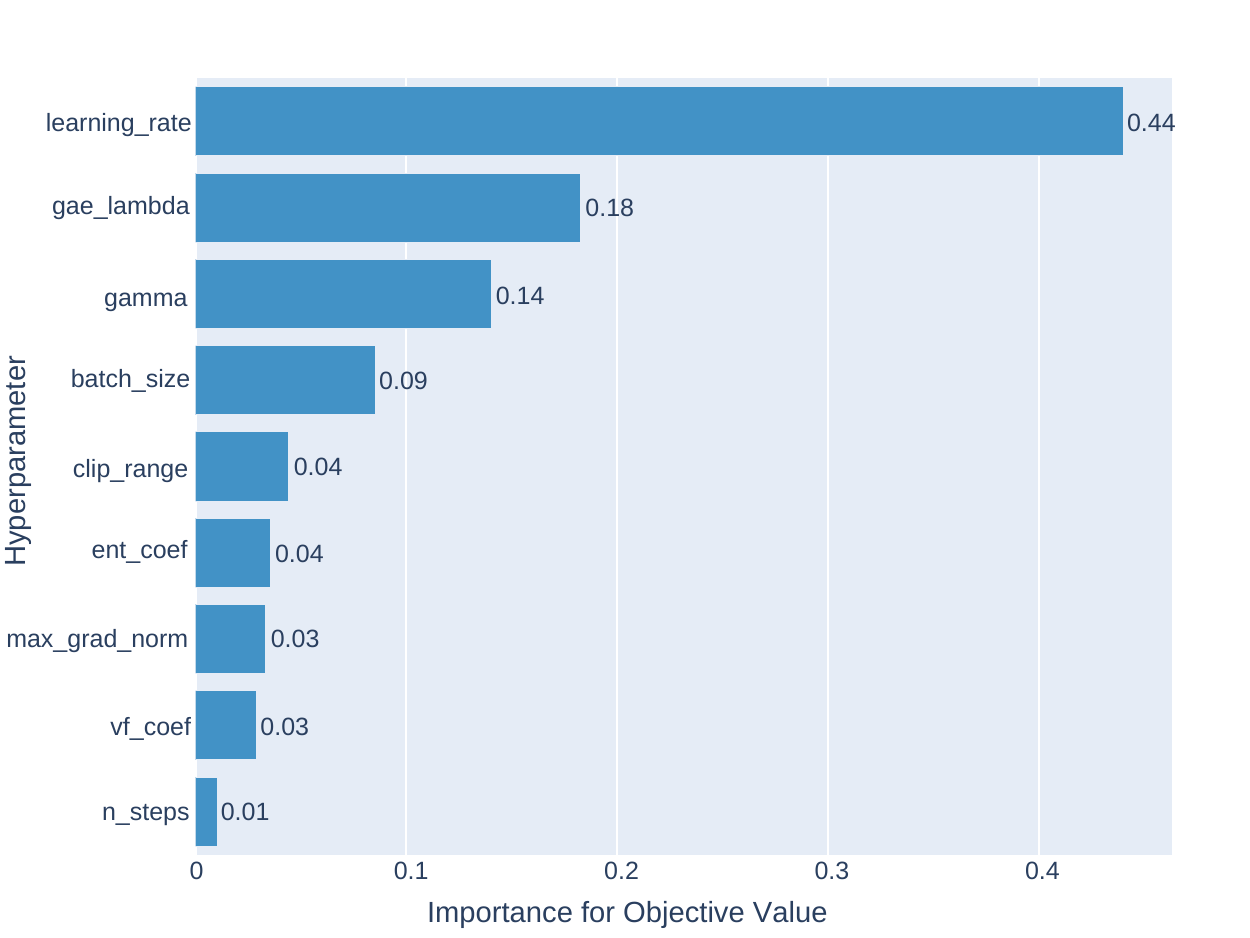}
        
    \end{minipage}%
    \begin{minipage}{0.5\textwidth}
        \centering
        \includegraphics[width=\linewidth]{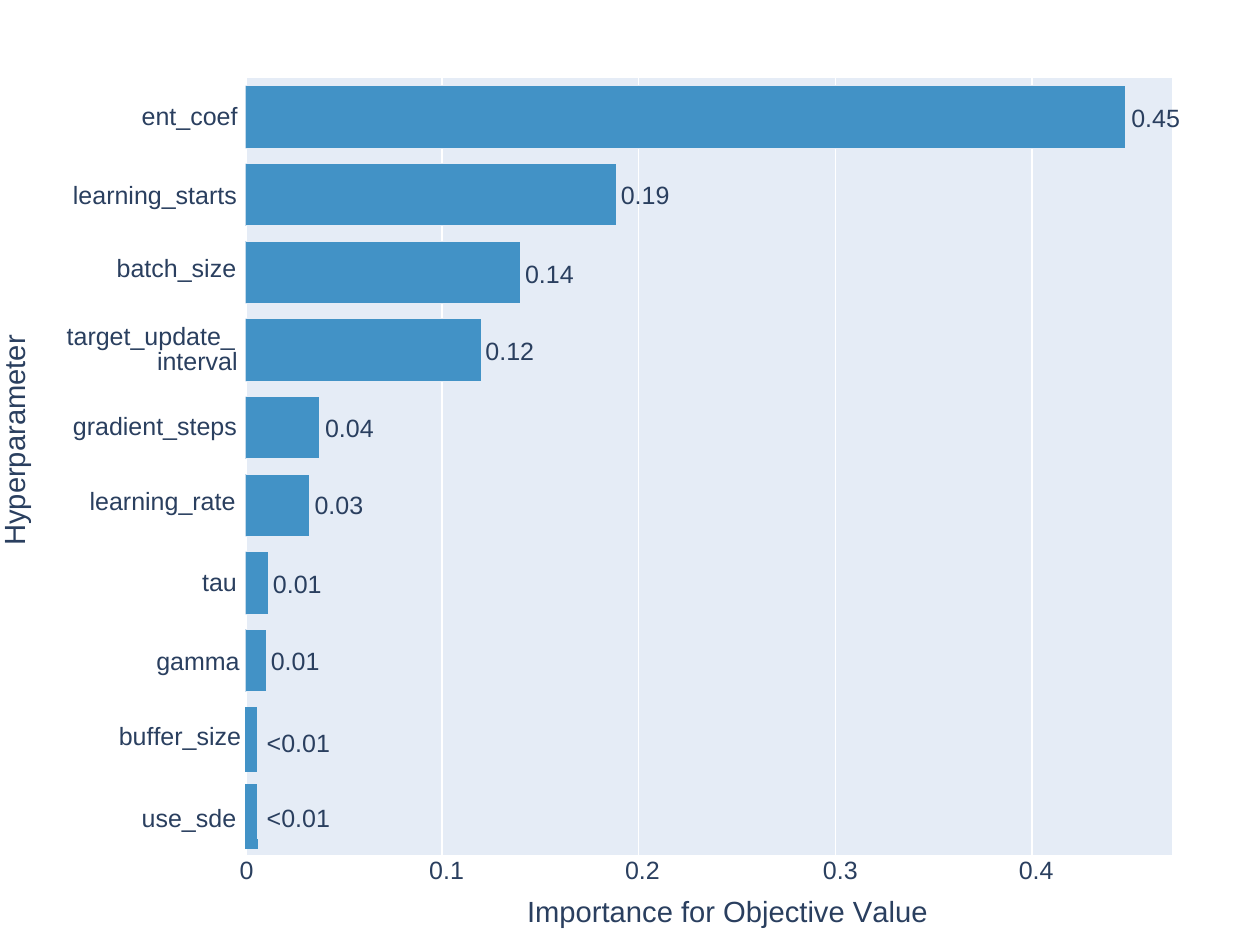}
        
    \end{minipage}
    \caption{Hyperparameters importance plot, for PPO \textbf{(left)} and SAC \textbf{(right)}.}\label{hip}
\end{figure}

The Hyperparameters Importance Plot, shown in Fig.~\ref{hip}, underscores the relative significance of each hyperparameter in influencing the success of the learning process. This plot is instrumental in identifying the hyperparameters that have the most substantial impact on the model's ability to learn efficiently.

A thorough examination of the SAC and PPO hyperparameters was conducted, detailing the range explored, and comparing the default values with the optimal values as determined by TPE. This comparison highlights TPE's nuanced approach to navigating the hyperparameter space and pinpointing the most effective configurations. The hyperparameters for PPO and SAC are listed in Table~\ref{PPO_h} and Table~\ref{SAC_h}, respectively. We selected default values based on recommendations from the Stable Baselines3 library \cite{stable-baselines3}.

\begin{table}[h]
\centering
\caption{Hyperparameters for Proximal Policy Optimization (PPO).}
\begin{tabular}{@{}lccc@{}}
\toprule
\textbf{Name} &\textbf{Default Value} & \textbf{Search space}  & \textbf{Best}\\
\midrule
learning\_rate &  0.0003 &  1e-5 - 1e-1 &  ~0.0153  \\
n\_steps & 2048 & 64 - 4096 &  559 \\
batch\_size & 64 & 16 - 256 &  193 \\
gamma  & 0.99 & 0.95 - 0.999 &  ~0.9657  \\
ent\_coef & 0.0 & 0.0 - 0.1 &  ~0.0548  \\
vf\_coef  & 0.5 & 0.2 - 1.0 &  ~0.3999  \\
max\_grad\_norm & 0.5 & 0.1 - 10.0 &  ~9.4229 \\
gae\_lambda & 0.95 & 0.8 - 0.99 &  ~0.8543 \\
clip\_range & 0.2 & 0.1 - 0.4 &  ~0.2865 \\
\hline
\end{tabular}
\label{PPO_h}
\end{table}

\begin{table}[h]
\centering
\caption{Hyperparameters for Soft Actor-Critic (SAC).}
\begin{tabular}{@{}lccc@{}}
\toprule
\textbf{Name} & \textbf{Default Value} & \textbf{Search space} & \textbf{Best}\\
\midrule
buffer\_size & 1000000 & 1000 - 1000000  &  79709\\
learning\_starts  & 1000 & 100 - 10000  &  7126\\
batch\_size & 256 & 16 - 256   &  104\\
tau & 0.005 & 0.001 - 0.1  &  ~0.034480\\
gamma & 0.99 & 0.9 - 0.999  &  ~0.920970\\
learning\_rate & 0.0003 & 1e-5 - 1e-1  &  ~0.000728\\
ent\_coef & 0.2 & 0.0 - 0.2  &  ~0.008345\\
target\_update\_interval & 1 & 1 - 100  &  40\\
gradient\_steps  & 1 & 1 - 20  &  10\\
use\_sde & False & True - False  &  True\\
\hline
\end{tabular}

\label{SAC_h}
\end{table}

\subsection{Enhanced Learning Efficiency}
\begin{figure}[h]
    \begin{minipage}{0.5\textwidth}
        \centering
        \includegraphics[width=\linewidth]{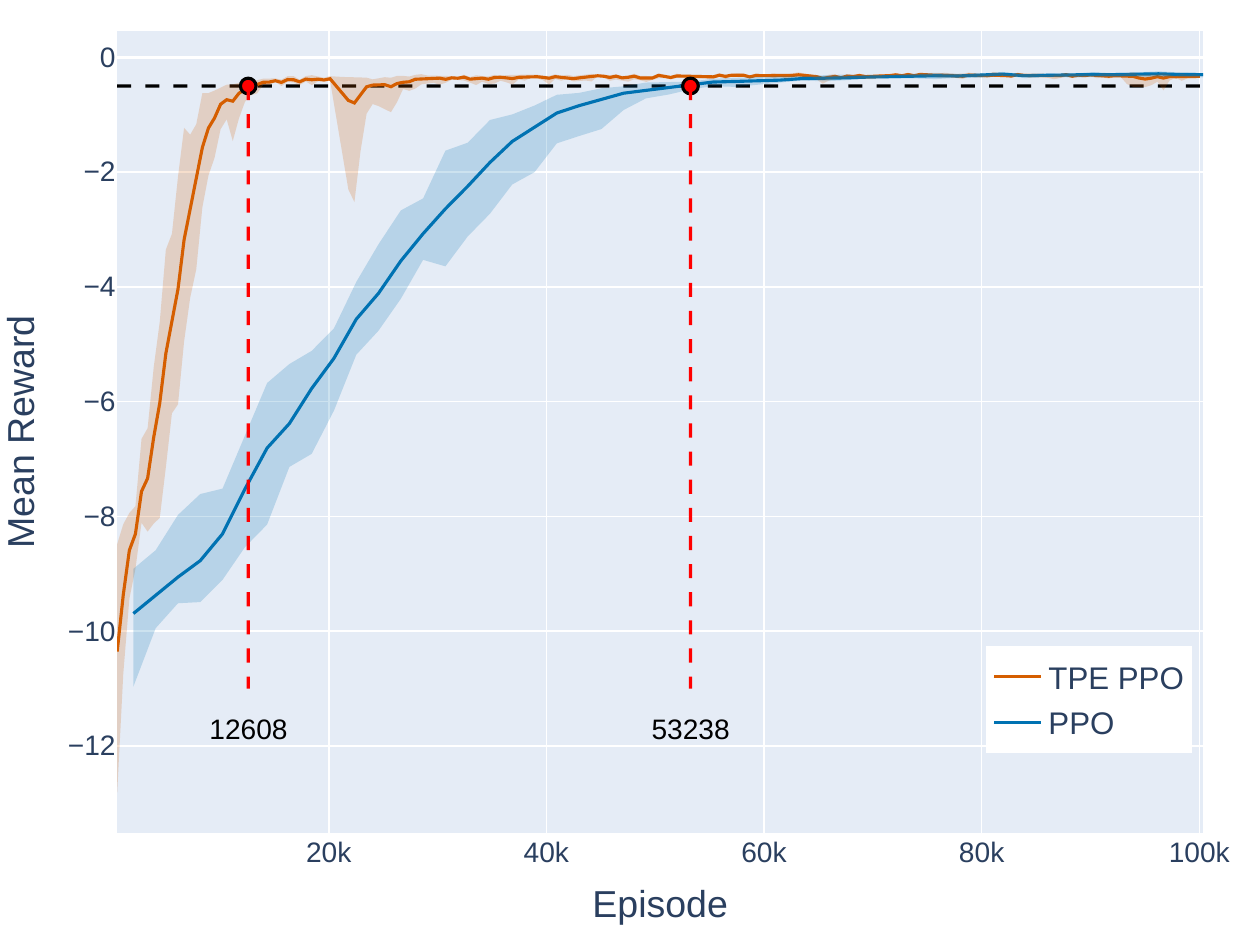}
        
    \end{minipage}%
    \begin{minipage}{0.5\textwidth}
        \centering
        \includegraphics[width=\linewidth]{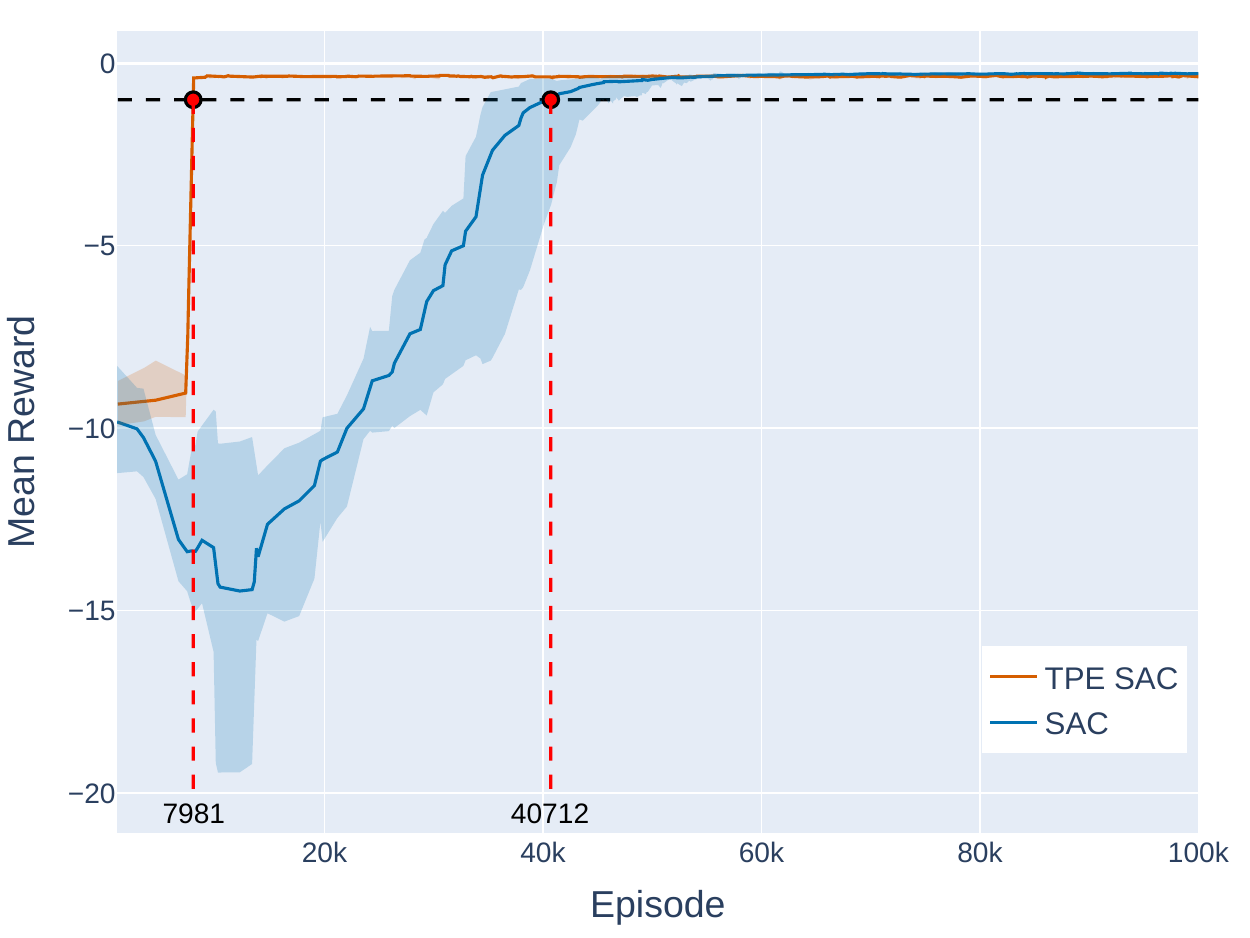}
        
    \end{minipage}
    \caption{Training Curves for PPO \textbf{(left)} and SAC \textbf{(right)} Before and After TPE Optimization showcases this improved learning trajectory, highlighting the efficiency gains in model training facilitated by TPE.}\label{TrainingCurves}
\end{figure}

The application of TPE markedly accelerated the learning process for both the SAC and PPO algorithms, as evidenced by the training curves (Fig.~\ref{TrainingCurves}). The training curves for each algorithm illustrated a faster ascent toward higher mean rewards per episode, indicative of an expedited convergence toward optimal policy decisions post-TPE optimization. With TPE, PPO converges to a reward within 95\% of the maximum payout 76.32\% faster than it could without it. This means that roughly 40K fewer training episodes are needed for PPO to function at its best. Furthermore, compared to no TPE, this improvement for SAC happens 80.39\% faster.

\subsection{Comparative Evaluation}
Upon evaluating the DRL models across a test set of 100,000 targets, Table~\ref{comparative_evaluation} showcases a distinct advantage in utilizing TPE-optimized hyperparameters. The results, detailed in Table~\ref{comparative_evaluation}, emphasize the superior performance of the optimized models, marking a significant improvement in success rates, and learning convergence speed for task achievements.

\begin{table}[h]
\centering
\caption{Success Rates of SAC and PPO Models Before and After Optimization}
\label{comparative_evaluation}
\begin{tabular}{@{}lccc@{}}
\hline

Model                      & \multicolumn{3}{c}{Success Rate (\%)} \\

\cmidrule(r){2-4}
 & 20K Episodes & 50K Episodes & 100K Episodes \\

\hline
SAC Before Optimization
 &3.22& 74.64 & 86.59
    \\
SAC After TPE Optimization & 
79.69& 85.12& 89.75
       \\
PPO Before Optimization & 
8.72&49.65& 87.29        
      \\
PPO After TPE Optimization
& 71.69&  83.93& 89.41
 
\\
\hline
\end{tabular}
\end{table}

\section{Conclusion}
This study demonstrates the benefits of using the TPE for hyperparameter optimization in DRL algorithms, SAC and PPO, for controlling robotic arms with 7-DOF. Our results indicate that TPE significantly improves both the learning efficiency and the performance of these algorithms, highlighted by the enhanced success rates and the accelerated convergence towards optimal rewards. Specifically, the application of TPE enabled PPO to achieve convergence to a reward within 95\% of the maximum reward 76.32\% faster than without TPE, necessitating about 40,630 fewer episodes of training for optimal performance, with similar improvements observed for SAC. These findings underscore the importance of precise hyperparameter tuning in the development of robust and efficient DRL models, particularly in the context of complex robotic tasks.

A notable challenge tackled in this research was the complexity of hyperparameter tuning, where TPE proved to be a valuable tool for optimizing the DRL algorithms' performance in robotic tasks.

Future research could broaden the scope by applying TPE to various DRL algorithms and robotic tasks such as pick-and-place, thereby unveiling more about its utility and limitations. Exploring other hyperparameter optimization methods could also contribute to refining the efficiency of DRL models further. Additionally, the practical application of these optimized models in real-world robotics scenarios will be crucial for making tangible advancements in the field. Our study paves the way for such explorations, indicating a promising direction for enhancing the performance and efficiency of DRL algorithms in complex, real-world tasks.

\subsubsection{\ackname}This work is supported by the University of Galway College of Science and Engineering Postgraduate Scholarship.

\bibliographystyle{ieeetr}
\bibliography{ref}
\end{document}